\documentclass[sigconf]{acmart}
\AtBeginDocument{%
  \providecommand\BibTeX{{%
    \normalfont B\kern-0.5em{\scshape i\kern-0.25em b}\kern-0.8em\TeX}}}

\copyrightyear{2022}
\acmYear{2022}
\setcopyright{acmcopyright}\acmConference[CIKM '22]{Proceedings of the 31st ACM
International Conference on Information and Knowledge Management}{October
17--21, 2022}{Atlanta, GA, USA}
\acmBooktitle{Proceedings of the 31st ACM International Conference on Information
and Knowledge Management (CIKM '22), October 17--21, 2022, Atlanta, GA, USA}
\acmPrice{15.00}
\acmDOI{10.1145/3511808.3557421}
\acmISBN{978-1-4503-9236-5/22/10}

\usepackage[linesnumbered,ruled,noend]{algorithm2e} 
\usepackage{multirow}
\usepackage{url}
\usepackage{bm}
\usepackage{amsmath,amsfonts}
\usepackage{textcomp}
\usepackage{xcolor}
\usepackage{amsthm}
\usepackage{subfigure}
\usepackage{paralist}
\usepackage{stfloats}
\usepackage{caption}
\usepackage{bbm}
\usepackage{siunitx}
\usepackage{setspace}
\usepackage{cancel}
\usepackage{hyperref}
\usepackage{balance}
\DeclareMathOperator*{\argmin}{arg\,min}
\begin{document}

\title{Prediction-based One-shot Dynamic Parking Pricing}

\author{Seoyoung Hong}
\affiliation{%
  \institution{Yonsei University}
  \city{Seoul}
  \country{Korea}}
\email{seoyoungh@yonsei.ac.kr}

\author{Heejoo Shin}
\affiliation{%
  \institution{University of California San Diego}
  \city{San Diego}
  \country{USA}}
\email{hes002@ucsd.edu}

\author{Jeongwhan Choi}
\affiliation{%
  \institution{Yonsei University}
  \city{Seoul}
  \country{Korea}}
\email{jeongwhan.choi@yonsei.ac.kr}

\author{Noseong Park}
\affiliation{%
  \institution{Yonsei University}
  \city{Seoul}
  \country{Korea}}
\email{noseong@yonsei.ac.kr}

\renewcommand{\shortauthors}{Seoyoung Hong, Heejoo Shin, Jeongwhan Choi, \& Noseong Park}

\begin{abstract}
Many U.S. metropolitan cities are notorious for their severe shortage of parking spots. To this end, we present a \emph{proactive} prediction-driven optimization framework to dynamically adjust parking prices. We use state-of-the-art deep learning technologies such as neural ordinary differential equations (NODEs) to design our future parking occupancy rate prediction model given historical occupancy rates and price information. Owing to the continuous and bijective characteristics of NODEs, in addition, we design a ``one-shot'' price optimization method given a pre-trained prediction model, which requires only one iteration to find the optimal solution. In other words, we optimize the price input to the pre-trained prediction model to achieve targeted occupancy rates in the parking blocks. We conduct experiments with the data collected in San Francisco and Seattle for years. Our prediction model shows the best accuracy in comparison with various temporal or spatio-temporal forecasting models. Our ``one-shot'' optimization method greatly outperforms other black-box and white-box search methods in terms of the search time and always returns the optimal price solution.
\end{abstract}

\ccsdesc[500]{Theory of computation~Mathematical optimization}
\ccsdesc[300]{Computing methodologies~Neural networks}

\keywords{dynamic pricing, parking occupancy prediction, price modeling, prediction-driven optimization}


\maketitle

\section{Introduction}
Maintaining an adequate parking occupancy rate is a highly essential but challenging problem in crowded metropolitan areas ~\cite{saharan2020efficient, shao2018parking, camero2018evolutionary}. In 2011-2013, for instance, San Francisco piloted SFpark, a demand-responsive parking pricing program adjusts parking prices based on observed occupancy rates.~\cite{pierce2013getting} If an observed occupancy rate is higher than a target occupancy rate, its price is increased. By adjusting prices in this manner, the program aimed to control parking occupancy rates around a target (ideal) rate of 70\%-80\%. SFpark's success has led to its adoption to other cities such as Los Angeles, Seattle, and Madrid~\cite{ghent2012express, pierce2013getting, lin2017survey}. They mainly focused on improving parking availability by keeping occupancy rates below the ideal rate, which is also of our utmost interest in this paper.


\begin{figure}[t]
    \setlength{\abovecaptionskip}{10pt}
    \setlength{\belowcaptionskip}{-4pt}
    \centering
    \subfigure[Black-box query-based prediction-driven optimization]{\includegraphics[width=\columnwidth]{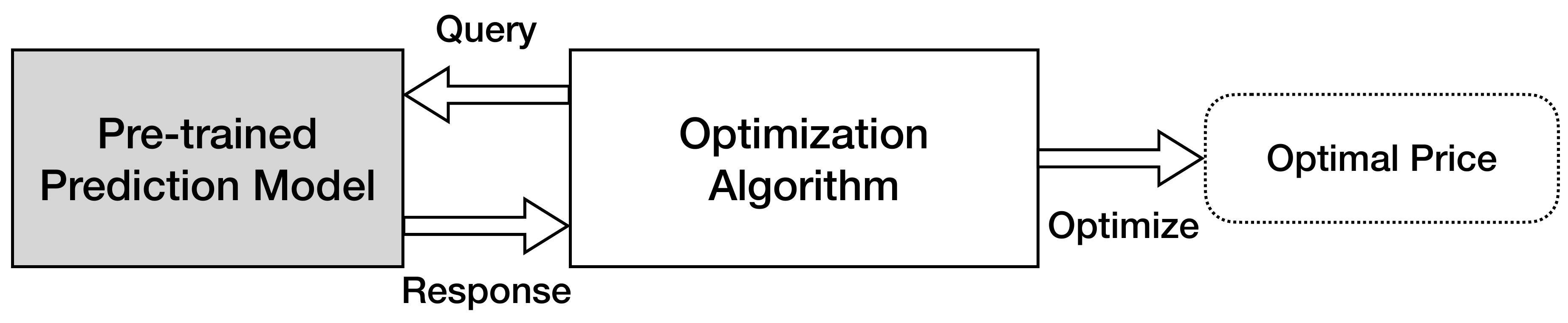}}
    \subfigure[Our proposed ``one-shot'', i.e., $\mathcal{O}(1)$-runtime, prediction-driven optimization]{\includegraphics[width=\columnwidth]{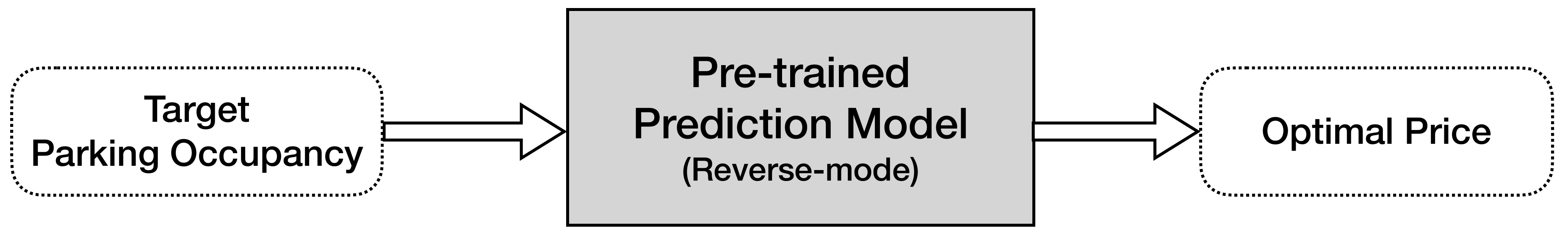}}
    \caption{Comparison of two paradigms of prediction-driven optimization. (a) The black-box query-based method typically requires a large number of queries until convergence of the solution. (b) As the name ``one-shot'' suggests, our method requires only one query to find the optimal solution. We can easily find the optimal price by solving the reverse-mode integral problem of the NODE layer in our model.}
    \label{fig:archi}
\end{figure}

In this work, we propose an advanced \emph{prediction-driven optimization} method to optimally adjust prices and achieve target on-street parking occupancy rates --- we conduct experiments for San Francisco and Seattle. Our approach differs from previous methods in the following sense: i) previous methods relied on observed parking demand to optimize price, i.e., \emph{reactive}, but we implement a prediction-based optimization model which predicts occupancy rates and optimizes prices based on those predictions, i.e., \emph{proactive}. ii) The price elasticity of demand varies depending on locations~\cite{pierce2013getting}, which implies that each block's demand may have a different degree of responsiveness to price changes. This is not considered in the existing approach, as it adjusts parking prices simultaneously for all blocks by fixed amounts. Our approach can change prices in a much more fine-grained manner, e.g., hourly and separately for each block. iii) We adjust the prices accurately, maximizing the influence of price adjustment.


\begin{figure}[t]
    \centering
     \subfigure[Overall workflow of how we predict the future occupancy rate]{\includegraphics[width=\columnwidth]{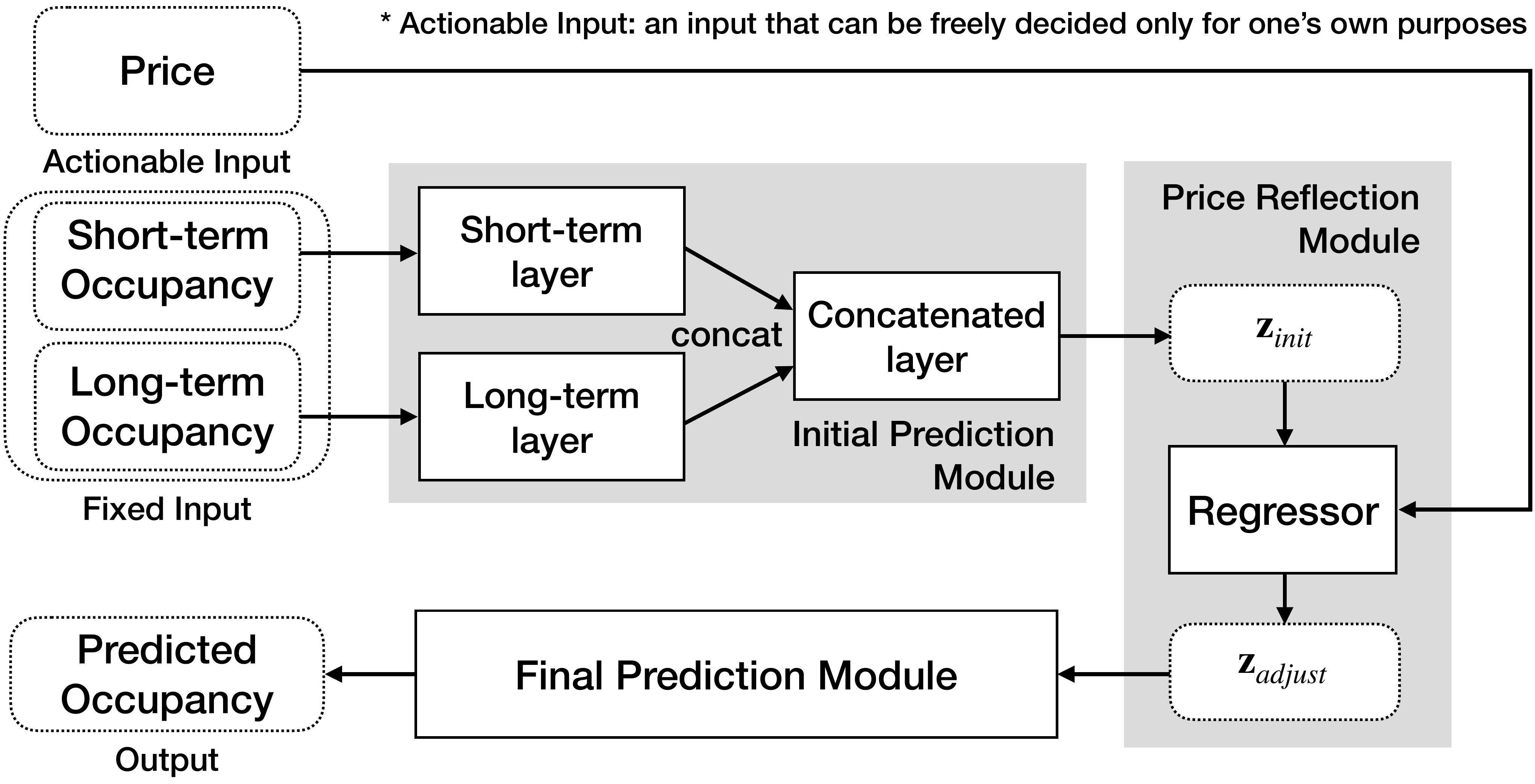}}
    \subfigure[Overall workflow of how we optimize the price]{\includegraphics[width=\columnwidth]{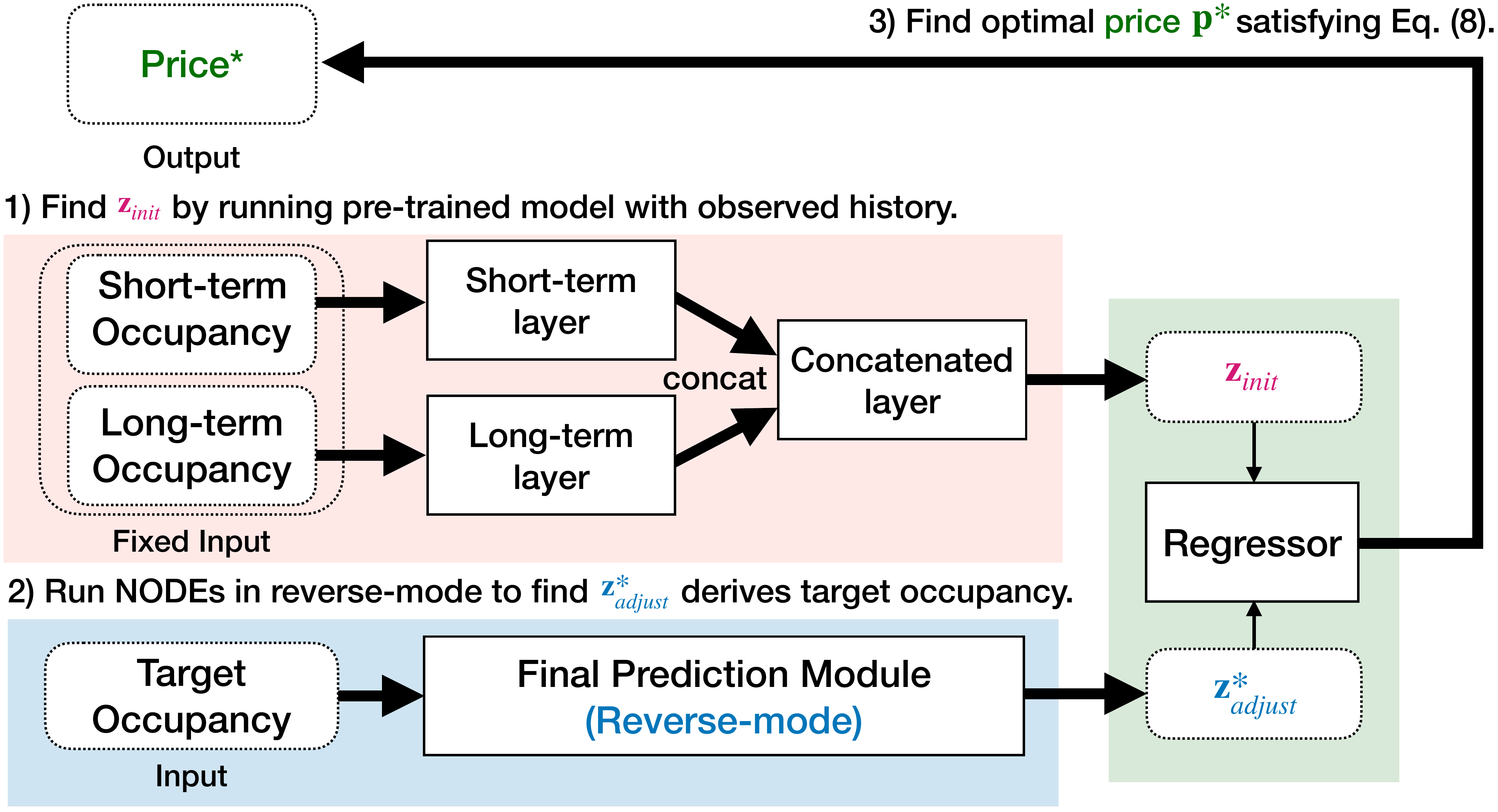}}
    \caption{The overall workflow of our prediction-driven dynamic price optimization. (a) Our predictive model consists of three parts: i) an initial prediction module with spatiotemporal processing to process short-term and long-term occupancy history, ii) a price reflection module to adjust the initial prediction with price information, and iii) a final prediction module to make the final predictions on the future occupancy rates of $N$ parking blocks. (b) Our price optimization process, which happens after training the prediction model, consists of three steps: i) given a pre-trained model and the short-term and long-term history, $\bm{z}_{init}$ is created as shown in the red box, ii)  given a pre-trained model and a target occupancy rate $\bm{y}^*$, we solve the reverse-mode integral problem to derive ${\bm{z}}^*_{adjust}$ as shown in the blue box, and iii) we find the optimal price $\bm{p}^*$ which minimizes the error between $\bm{z}_{init}$ and ${\bm{z}}^*_{adjust}$ in the green box (cf. Eq.~\eqref{eq:opt}).}
    \label{fig:archi1}
\end{figure}

 Given a set of parking blocks, denoted $\{b_i\}_{i=1}^{N}$, let $\bm{y}^* \in [0,1]^N$ be an $N$-dimensional vector which contains target occupancy rates for $N$ parking blocks. Our method finds $\bm{p}^* \in [p_{min},p_{max}]^N$, where $p_{max}$/$p_{min}$ denotes the maximum/minimum price, which makes the pre-trained occupancy prediction model output $\bm{y}^*$ (cf. Fig.~\ref{fig:archi} (b)). In this research, it is important to design a high-accuracy prediction model since it is imperative to guarantee the quality of price optimization. The most important technical concern is, however, how to integrate the prediction model and the optimization method because an ill-integrated prediction-driven optimization does not return a solution in time  ~\cite{An:2016:MFM:2939672.2939726,An:2017:DFA:3055535.3041217,li2021largescale,doi:10.1137/1.9781611976700.80}. The simplest method is to use \emph{black-box queries}, where the optimization algorithm queries the prediction model about the quality of a solution (cf. Fig.~\ref{fig:archi} (a)). This black-box method typically requires a non-trivial number of queries. A better approach is the \emph{gradient-based white-box method}. Gradients are created from the internals of the prediction model to update the input feature (i.e., the parking prices in our case). This approach achieves a shorter search time than that of the black-box method. However, this also requires multiple iterations to achieve a reliable solution. 

In this work, we propose yet another ``one-shot'' white-box search method that can find the optimal price solution with only \emph{one query} to the prediction model, owing to the continuous and bijective characteristic of neural ordinary differential equations (NODEs~~\cite{NIPS2018_7892}). Fig.~\ref{fig:archi1} shows our model design. Since we solve two coupled problems in this paper, the occupancy rate prediction and the price optimization, Figs.~\ref{fig:archi1} (a) and (b) show the two workflows of a) how we forecast the future occupancy rates and b) how we optimize the prices, respectively. The final prediction module in our model, which consists of NODEs, is continuous and bijective, and our one-shot prediction-driven optimization process in Fig.~\ref{fig:archi1} (b) is possible.

Our prediction model consists of three modules as shown in Fig.~\ref{fig:archi1} (a): i) the initial prediction module, ii) the price reflection module, and iii) the final prediction module. The initial prediction module is further decomposed into three layers, depending on the type of processed information. The short-term layer processes the occupancy rates during $K$ recent periods and the long-term layer processes the past occupancy rates older than the short-term information, e.g., a week ago. The concatenated processing layer combines the two hidden representations, one by the short-term layer and the other by the long-term layer, to produce the initial predictions on the future occupancy rates of all those $N$ parking blocks. Up to this moment, however, we do not consider the price information yet, and it is the second price reflection module that processes the price information to enhance the initial occupancy predictions. The final prediction module fine-tunes the predictions to increase the model accuracy. The continuous and bijective characteristics of the final prediction module enables the one-shot price optimization.

The detailed price optimization process is depicted in Fig.~\ref{fig:archi1} (b). We first train the predictive model and then give the observed short/long-term occupancy information and the target occupancy rates $\bm{y}^*$, we decide the best prices $\bm{p}^*$ which yield the target rates. The red box of Fig.~\ref{fig:archi1} (b) is first processed with the observed short/long-term input to derive the initial prediction $\bm{z}_{init}$ and the blue box is processed to derive ${\bm{z}}^*_{adjust}$. We note that at this moment, the input and output of the final prediction module are opposite in comparison with Fig.~\ref{fig:archi1} (a), which is theoretically possible due to its continuous and bijective properties. Then, we calculate $\bm{p}^*$ which matches $\bm{z}_{init}$ to ${\bm{z}}^*_{adjust}$.

We conduct experiments on two datasets collected in San Francisco and Seattle for multiple years. Our prediction model outperforms various baselines, and our one-shot optimization can very quickly find the optimal prices for most of the parking blocks during the testing period, whereas other black-box and white-box methods fail to do so. In addition, our visualization results intuitively show that our proposed method is effective in adjusting the parking occupancy rates to on or below the ideal occupancy rate. Our contributions can be summarized as follows:

\begin{enumerate}
    \item We design a sophisticated future parking occupancy rate prediction model based on NODEs.
    \item We design a novel one-shot price optimization, owing to the continuous and bijective characteristics of NODEs.
    \item In our experiments with two real-world datasets, our prediction model outperforms many existing temporal and spatiotemporal models, and our one-shot optimization finds better solutions in several orders of magnitude faster in comparison with other prediction-driven optimization paradigms.
\end{enumerate}

\section{Related Work and Preliminaries}

\subsection{Neural Ordinary Differential Equations}\label{sec:node}
NODEs solve the following integral problem to calculate the last hidden vector $\bm{z}(T)$ from the initial vector $\bm{z}(0)$~\cite{NIPS2018_7892}:
\begin{align}\label{eq:node}
    \bm{z}(T) = \bm{z}(0) + \int_{0}^{T}f(\bm{z}(t);\bm{\theta}_f)dt,
\end{align}where $f(\bm{z}(t);\bm{\theta}_f)$, which we call \emph{ODE function}, is a neural network to approximate $\dot{\bm{z}} \stackrel{\text{def}}{=} \frac{d \bm{z}(t)}{d t}$. To solve the integral problem, NODEs rely on ODE solvers, e.g., the explicit Euler method, the Dormand--Prince (DOPRI) method, and so forth~\cite{DORMAND198019}.

Let $\phi_t : \mathbb{R}^{\dim(\bm{z}(0))} \rightarrow \mathbb{R}^{\dim(\bm{z}(T))}$ be a mapping from $t=0$ to $t=T$ created by an ODE after solving the integral problem. It is well-known that $\phi_t$ becomes a homeomorphic mapping~\cite{NIPS2019_8577,massaroli2020dissecting}: $\phi_t$ is continuous and bijective and $\phi_t^{-1}$ is also continuous for all $t \in [0,1]$. It was shown that the homeomorphic characteristic increases the robustness of NODEs to scarce input~\cite{kim2021oct,choi2021ltocf}. We conjecture that NODEs are, for the same reason, also suitable for our occupancy prediction, since we cannot feed abundant information to our model due to the difficulty of collecting other auxiliary data. Occupancy prediction is an information-scarce task and therefore, the robustness of the model is crucial for our task.

\paragraph{Reverse-mode integral problem} In addition, the bijective characteristic makes our prediction optimization process easier than other methods. For instance, let $\bm{z}^*(T)$ be the preferred output that we want to see. Our price optimization corresponds to finding the $\bm{z}^*(0)$ that leads to $\bm{z}^*(T)$. In the case of NODEs, for finding $\bm{z}^*(0)$, we can solve the reverse-mode integral problem, i.e., $\bm{z}^*(0) = \bm{z}^*(T) - \int_{0}^{T}f(\bm{z}(t);\bm{\theta}_f)dt$, and $\bm{z}^*(0)$ is unique. 

\paragraph{Adjoint sensitivity method} Instead of the backpropagation method, the adjoint sensitivity method is used to train NODEs for its efficiency and theoretical correctness~\cite{NIPS2018_7892}. After letting $\bm{a}_{\bm{z}}(t) = \frac{d \mathcal{L}}{d \bm{z}(t)}$ for a task-specific loss $\mathcal{L}$, it calculates the gradient of loss w.r.t model parameters with another reverse-mode integral as follows:\begin{align*}\nabla_{\bm{\theta}_f} \mathcal{L} = \frac{d \mathcal{L}}{d \bm{\theta}_f} = -\int_{t_m}^{t_0} \bm{a}_{\bm{z}}(t)^{\mathtt{T}} \frac{\partial f(\bm{z}(t);\bm{\theta}_f)}{\partial \bm{\theta}_f} dt.\end{align*}

$\nabla_{\bm{z}(0)} \mathcal{L}$ can also be calculated similarly and we can propagate the gradient backward to layers before the ODE if any. It is worth mentioning that the space complexity of the adjoint sensitivity method is $\mathcal{O}(1)$ whereas using the backpropagation to train NODEs has a space complexity proportional to the number of DOPRI steps. The time complexity of the adjoint sensitivity method is similar, or slightly more efficient than that of backpropagation. Therefore, we can train NODEs efficiently.



\subsection{Parking Occupancy Prediction}


Parking management in metropolitan areas has long been a matter of interest and researched from diverse perspectives since the 1970s. In recent years, there have been many studies on predicting parking availability, of which several have focused on the SFpark data. Two studies compared the efficacy of regression trees and various regression methods for predicting San Francisco's parking occupancy rates~\cite{zheng2015parking, simhon2017smart}. Other studies have been conducted on other data, such as Santander, Spain~\cite{vlahogianni2016real}; Birmingham, UK~\cite{camero2018evolutionary}; Melbourne, Australia~\cite{shao2018parking}. Traditional machine learning algorithms have been extended for processing spatiotemporal data, e.g., traffic forecasting~\cite{bing2018stgcn, li2018dcrnn_traffic, NEURIPS2020_ce1aad92,choi2022STGNCDE}. According to the survey~\cite{jiang2021graph}, two recent researches~\cite{zhang2020semi, jiang2021graph} used a spatiotemporal model to predict the parking occupancy rates of Beijing and Shenzhen in China. These models are fine-grained and use different features and datasets. Most importantly, they do not consider the price information and their models are designed without considering the price optimization. We instead compare our model with spatiotemporal models. 

\subsection{Parking Price Optimization}


The parking price optimization problem can be formulated as adjusting prices to minimize the error between predicted and target occupancy rates. In~\cite{fabusuyi2018rethinking}, they introduced a predictive optimization strategy that enables proactive parking pricing rather than a reactive approach based on observed parking rates. By implementing a regression-based prediction model considering price elasticity, they optimized parking prices. However, their model is a simple regression for which prediction-driven optimization is technically straightforward. In~\cite{saharan2020efficient}, they compared various machine learning prediction models and utilized them to produce occupancy-based parking prices for Seattle city's on-street parking system.

These papers have adopted prediction-based optimization, formulating pricing schemes based on predicted occupancy rates (as in ours). However, our method is novel in the following aspects:
\begin{compactenum}
    \item Most of the previous work implemented predictive models with simple machine learning approaches such as linear regression. Our model uses the latest and advanced deep learning approaches, such as NODEs.
    \item Traditional optimization methods, such as greedy, gradient-based optimization, and so forth, were used for previous papers. We greatly reduce the running time of our method using the novel one-shot optimization which relies on the continuous and bijective nature of NODEs.
\end{compactenum}

\section{Occupancy Rate Prediction}

\subsection{Overall Architecture}
As shown in Fig.~\ref{fig:archi1}, our proposed method consists of three modules: Firstly, the initial prediction module with spatiotemporal processing can be further decomposed into the following three layers:
\begin{compactenum}
    \item The short-term layer processes the short-term occupancy rate information. This layer only considers $K$ recent occupancy rates and produces the hidden representation matrix $\bm{H}_{short} \in \mathbb{R}^{N \times \dim(\bm{H}_{short})}$.
    \item The long-term layer reads the long-term occupancy rate information (i.e., the mean occupancy rates on the same day of the last week/last two weeks/last month/last two months and so forth). Since the long-term occupancy rate has irregular time intervals, we found that the fully-connected layer is a reasonable design for this layer. This layer produces the hidden representation matrix $\bm{H}_{long}\in \mathbb{R}^{N \times \dim(\bm{H}_{long})}$.
    \item Given the concatenated hidden representation matrix $\bm{H}_{short} \oplus \bm{H}_{long}$, where $\oplus$ means the horizontal concatenation, the concatenated processing layer produces the initial future occupancy rate prediction $\bm{z}_{init} \in \mathbb{R}^{N \times 1}$.
\end{compactenum}

Secondly, the price reflection module adjusts $\bm{z}_{init}$ created by the initial prediction module since it does not consider the price information yet. Given $\bm{z}_{init}$, this module adjusts it to $\bm{z}_{adjust} \in \mathbb{R}^{N \times 1}$ by considering the inputted price information. 
    
Lastly, the final prediction module evolves $\bm{z}_{adjust}$ to the final prediction $\hat{\bm{y}} \in [0,1]^{N \times 1}$. We use a series of NODEs since we found that a single NODE does not return reliable predictions.

\subsection{Initial Prediction Module}

We note that in this module, all parking blocks' short-term and long-term occupancy information is processed altogether. Therefore, our model is able to consider, when predicting for a parking block, not only its own historical information but also other related parking blocks' historical information, i.e., spatiotemporal processing.

\paragraph{Short-term history layer} Given a set of $K$ recent short-term occupancy rates, denoted $\{\bm{s}_i\}_{i=1}^K$, where $\bm{s}_i \in [0,1]^N$, $N$ refers to the number of parking blocks, and $\bm{s}_K$ represents the most recent record, we extract the hidden representation of the short-term history. We note that $\{\bm{s}_i\}_{i=1}^K$ constitutes a time-series sample for which any type of time-series processing technique can be applied. This layer is to use the following NODE where $\bm{H}(0)$ is created from $\{\bm{s}_i\}_{i=1}^K$:
\begin{align}\label{eq:short}
    \bm{H}_{short} = \bm{H}(0) + \int_0^1 f(\bm{H}(t);\bm{\theta}_f) dt,
\end{align}

The ODE function $f:\mathbb{R}^{N \times \dim(\bm{H}_{short})} \rightarrow \mathbb{R}^{N \times \dim(\bm{H}_{short})}$ is defined as follows:
\begin{align*}\begin{split}
f(\bm{H}(t);\bm{\theta}_f) =& \psi(FC_{N \times \dim(\bm{H}_{short}) \rightarrow N \times \dim(\bm{H}_{short})}(\bm{E}_1)),\\
\bm{E}_1 =& \sigma(FC_{N \times \dim(\bm{H}_{short}) \rightarrow N \times \dim(\bm{H}_{short})}(\bm{E}_0)),\\
\bm{E}_0 =& \sigma(FC_{N \times \dim(\bm{H}_{short}) \rightarrow N \times \dim(\bm{H}_{short})}(\bm{H}(t))),
\end{split}\end{align*} where $\sigma$ is a rectified linear unit, $\psi$ is a hyperbolic tangent, $\bm{\theta}_f$ refers to the parameters of the three fully-connected layers. Note that this ODE layer's hidden size is the same as the input size $\dim(\bm{H}_{short})$. 

\paragraph{Long-term history layer} Let $\{\bm{l}_i\}_{i=1}^{L}$, where $\bm{l}_i \in [0,1]^{N}$, be a set of the long-term occupancy rate information. We have 12 types of the long-term historical information for each parking block, i.e., $L=12$, which includes the past occupancy rates at different time points. We use the following fully-connected layer to process the long-term history $\{\bm{l}_i\}_{i=1}^{L}$ since they do not clearly constitute a time-series sample:
\begin{align}
    \bm{H}_{long} = \texttt{FC}_{N \times L \rightarrow N \times L}(\oplus_{i=1}^{L} \bm{l}_i^\intercal),
\end{align}where $\texttt{FC}_{N \times L \rightarrow N \times L}$ means the fully-connected layer with an input size (i.e., the dimensionality of input matrix) of $N \times L$ to an output size of $N \times L$.

\paragraph{Concatenated processing layer} We horizontally combine $\bm{H}_{short}$ and $\bm{H}_{long}$ to produce the initial prediction $\bm{z}_{init}$. We evolve $\bm{z}_{init}$ from $\bm{H}_{short}$ by referring to $\bm{H}_{long}$. In this layer, we have one more augmented NODE layer as follows:
\begin{align}
    \bm{z}_{init} = \bm{c}(1) = \bm{c}(0) + \int_0^1 m(\bm{c}(t), \bm{H}_{long};\bm{\theta}_m) dt,
\end{align} where $\bm{c}(0) = \texttt{FC}_{N \times \dim(\bm{H}_{short}) \rightarrow N \times 1}(\bm{H}_{short})$. We note that after this processing, $\bm{z}_{init} \in \mathbb{R}^{N \times 1}$. For reducing the overall computational overhead of our method, we early make the initial prediction. 

The ODE function $m:\mathbb{R}^{N\times(L+1)} \rightarrow \mathbb{R}^{N \times 1}$ is defined as follows:
\begin{align*}\begin{split}
m(\bm{c}(t), \bm{h}_{long};\bm{\theta}_m) =& \psi(FC_{N \times 1 \rightarrow N \times 1}(\bm{u}_1)),\\ \bm{u}_1 =& \sigma(FC_{N \times 1 \rightarrow N \times 1}(\bm{u}_0)),\\
\bm{u}_0 =& \sigma(FC_{N\times(L+1) \rightarrow N \times 1}(\bm{c}(t) \oplus \bm{h}_{long})),
\end{split}\end{align*}where $\bm{\theta}_m$ refers to the parameters of the three fully-connected layers. In particular, this type of NODEs is called as augmented NODEs, which can be written as follows for our case:
\begin{align}
\frac{d}{dt}{\begin{bmatrix}
  \bm{c}(t) \\
  \bm{h}_{long} \\
  \end{bmatrix}\!} = {\begin{bmatrix}
  m(\bm{c}(t), \bm{h}_{long};\bm{\theta}_m) \\
  0\\
  \end{bmatrix}\!}.
\end{align}

\subsection{Price Reflection Module}
The above initial prediction module does not consider one of the most important factors in forecasting the future occupancy, which is the price information. We reflect the price information $\bm{p} \in [p_{min},p_{max}]^{N}$ into $\bm{z}_{init}$ and create $\bm{z}_{adjust}$ in this module. We use the following regressor for this module:
\begin{align}
    \bm{z}_{adjust} = \bm{z}_{init} - \Big((\bm{c}\odot\bm{p})^\intercal + \bm{b}\Big),
\label{eq:price}
\end{align}where $\bm{c} \in [0,\infty]^N$ is a coefficient vector that represents the demand elasticity on price changes, i.e. how much the occupancy rate reacts to the price. $\odot$ means the element-wise product. $\bm{b} \in \mathbb{R}^{N \times 1}$ is a bias (column) vector.

\subsection{Final Prediction Module}
Given the adjusted prediction $\bm{z}_{adjust} \in \mathbb{R}^{N \times 1}$, there are multiple NODE layers to evolve it to the future occupancy prediction $\hat{\bm{y}}$. Since we found that only one NODE layer is not enough for this purpose, we adopt the following $M$ NODE layers:
\begin{align}\begin{split}
    \bm{y}_1(1) &= \bm{z}_{adjust} + \int_0^1 j_1(\bm{y}_1(t);\bm{\theta}_{j_1}) dt,\\
    \bm{y}_i(1) &= \bm{y}_{i-1}(1) + \int_0^1 j_i(\bm{y}_i(t);\bm{\theta}_{j_i}) dt, 1<i<M,\\
    \bm{y}_M(1) &= \bm{y}_{M-1}(1) + \int_0^1 j_M(\bm{y}_M(t);\bm{\theta}_{j_M}) dt,
\end{split}\label{eq:final}\end{align}where the future occupancy prediction $\hat{\bm{y}} = \bm{y}_M(1)$. 

We use the following identical architecture for $j_i:\mathbb{R}^{N \times 1} \rightarrow \mathbb{R}^{N \times 1}$, for all $1\leq i \leq M$, but their parameters $\bm{\theta}_{j_i}$ are different:
\begin{align*}\begin{split}
j_i(\bm{y}_i(t);\bm{\theta}_{j_i}) =& \psi(FC_{N \times 1 \rightarrow N \times 1}(\bm{o}_1)),\\
\bm{o}_1 =& \sigma(FC_{N \times 1 \rightarrow N \times 1}(\bm{o}_0)),\\
\bm{o}_0 =& \sigma(FC_{N \times 1 \rightarrow N \times 1}(\bm{y}_i(t))).
\end{split}\end{align*}

\begin{algorithm}[t]
\RestyleAlgo{ruled}
\caption{How to train our model}\label{alg:train}
\KwIn{Training data $D_{train}$, Validating data $D_{val}$, Maximum iteration number $max\_iter$}
Initialize all parameters, denoted $\theta_{all}$;

$iter \gets 0$;

\While {$iter < max\_iter$}{
    Train all parameters $\theta_{all}$ with the loss $\mathcal{L}$;
    
    Validate and update the best parameters\;
    
    $iter \gets iter + 1$;
}
\Return the selected best parameters;
\end{algorithm}

\subsection{Training Algorithm}
We use the adjoint method \cite{NIPS2018_7892} to train each NODE layer in our model, which requires a memory of $\mathcal{O}(1)$ and a time of $\mathcal{O}(\delta)$ where $\delta$ is the (average) step-size of an underlying ODE solver. In Alg.~\eqref{alg:train}, we show our training algorithm. We use the following mean squared error loss with the training data $D_{train}$:
\begin{align*}
    \mathcal{L} = \frac{\sum_{i=1}^{|D_{train}|}\|\bm{y}_i - \hat{\bm{y}}_i\|^2_2}{|D_{train}|} + w\|\theta_{all}\|^2_2,
    \label{eq:loss}
\end{align*}where $\bm{y}_i$ and $\hat{\bm{y}}_i$ mean the ground-truth and predicted occupancy rate of the $i$-th training sample, respectively. $w$ is the coefficient of the $L^2$ regularization term. $\theta_{all}$ denotes the super set of the all parameters in our model.

\paragraph{Well-posedness of Training}
The well-posedness\footnote{A well-posed problem means i) its solution uniquely exists, and ii) its solution continuously changes as input data changes.} of NODEs was already proved in \cite[Theorem 1.3]{lyons2004differential} under the mild condition of the Lipschitz continuity. We show that training our NODE layers is also a well-posed problem. Almost all activations, such as ReLU, Leaky ReLU, SoftPlus, Tanh, Sigmoid, ArcTan, and Softsign, have a Lipschitz constant of 1. Other common neural network layers, such as dropout, batch normalization and other pooling methods, have explicit Lipschitz constant values. Therefore, the Lipschitz continuity of $f$, $m$ and $j_i$ for all $i$ can be fulfilled in our case, making it is a well-posed training problem. Our training algorithm solves a well-posed problem so its training process is stable in practice.


\section{One-shot Price Optimization}
We describe our proposed dynamic pricing method. The key algorithm is the following ``one-shot'' price optimization method which finds a solution in one iteration. For this step, we pre-train the prediction model and all its parameters are considered as constants during this step.

Given the short-term history information $\{\bm{s}_i\}_{i=1}^K$, the long-term history information $\{\bm{l}_i\}_{i=1}^{L}$, and the target occupancy rates $\bm{y}^*$, we want to find the optimal prices $\bm{p}^*$ that leads to $\bm{y}^*$ as follows:
\begin{align}\begin{split}
    \argmin_{\bm{p}^*}\;\;\; &\frac{\|\hat{\bm{y}} - \bm{y}^*\|_1}{N},\\
    \textrm{subject to}\;\;\; & \bm{p}_{min} \leq \bm{p}^* \leq \bm{p}_{max},\\
    & \hat{\bm{y}} = \xi(\bm{p}^*, \{\bm{s}_i\}_{i=1}^K, \{\bm{l}_i\}_{i=1}^{L}; \bm{\theta}_{\xi}),
\end{split}\end{align}where $\xi$ is our pre-trained occupancy prediction model, and $\bm{\theta}_{\xi}$ is its set of parameters. This becomes a complicated non-linear resource allocation problem, whose polynomial-time solver is unknown, if $\xi$ is a general deep neural network~\cite{HOCHBAUM1995103,5374378}. For those reasons, people typically rely on the Karush--Kuhn--Tucker (KKT) condition, when it is possible to calculate the derivative of objective, which teaches us the necessary condition of optimality to find a reasonable solution, but this method sometimes converges to saddle points that are not good enough~\cite{10.5555/1355334}. In our case, however, we use the following one-shot method due to our special design suitable for solving the problem:
\begin{compactenum}
    \item We feed the given short-term and the long-term history information and derive $\bm{z}_{init}$ from the initial prediction module;
    \item Let $\bm{y}^*$ is the target occupancy rates that we want to achieve. We then solve the following series of reverse-mode integral problems to derive ${\bm{z}}^*_{adjust}$:
\begin{align*}\begin{split}
    \bm{y}_{M-1}(1) &= \bm{y}^* - \int_0^1 j_M(\bm{y}_M(t);\bm{\theta}_{j_M}) dt,\\
    \bm{y}_{i-1}(1) &= \bm{y}_{i}(1) - \int_0^1 j_i(\bm{y}_i(t);\bm{\theta}_{j_i}) dt, 1<i<M,\\
    {\bm{z}}^*_{adjust} &= \bm{y}_1(1) - \int_0^1 j_1(\bm{y}_1(t);\bm{\theta}_{j_1}) dt;
\end{split}\end{align*}
    \item The optimal price vector $\bm{p}^*$ can be calculated as follows:
    \begin{align} \label{eq:opt}
    \bm{p}^* = \frac{\bm{z}_{init} - {\bm{z}}^*_{adjust} - \bm{b}}{\bm{c}}.
    \end{align}
\end{compactenum}

We note that we query the initial prediction module and solve each of the integral problems exactly once, i.e., a constant complexity of $\mathcal{O}(1)$ given a fixed number $M$ of NODE layers, which is theoretically the minimum complexity that we can achieve (since optimization with zero queries is infeasible). Note $M$ does not vary during operation but is fixed given a prediction model.

\begin{table}[t]
\renewcommand{\arraystretch}{0.85}
\caption{The size of the training/testing datasets}
\label{dataset}
\begin{tabular}{l|l|cc}
\toprule
                                                    &       & \multicolumn{1}{l}{\textbf{Training Dataset}} & \multicolumn{1}{l}{\textbf{Test Dataset}} \\ \midrule
\multicolumn{1}{c|}{\multirow{3}{*}{San Francisco}} & $K=1$ & 407,008                              & 183,280                          \\
\multicolumn{1}{c|}{}                               & $K=2$ & 356,132                              & 160,370                          \\
\multicolumn{1}{c|}{}                               & $K=3$ & 305,256                              & 137,460                          \\ \midrule
\multirow{3}{*}{Seattle}                            & $K=1$ & 75,264                               & 31,360                           \\
                                                    & $K=2$ & 65,856                               & 27,440                           \\
                                                    & $K=3$ & 56,448                               & 23,520                           \\
\bottomrule
\end{tabular}
\end{table}

\begin{figure}[t]
    \centering
    \includegraphics[width=\columnwidth]{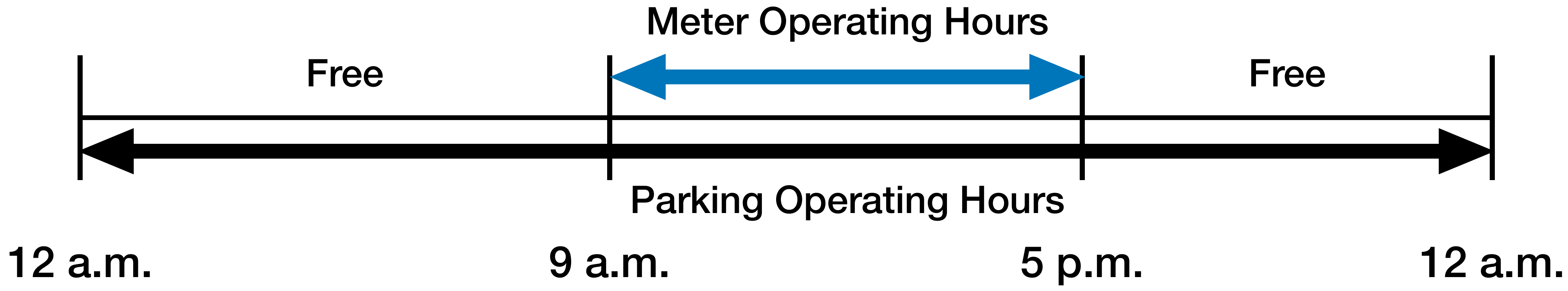}
    \setlength{\abovecaptionskip}{0pt}
    \setlength{\belowcaptionskip}{-4pt}
    \caption{The visualization of parking operating hours and meter operating hours of San Francisco}
    \label{meter_oper}
\end{figure}

\section{Experiments on Prediction}
Our software and hardware environments are as follows:  \textsc{Ubuntu} 18.04 LTS, \textsc{Python} 3.7.10, \textsc{PyTorch} 1.6.0, \textsc{TORCHDIFFEQ} 0.2.2, \textsc{TORCHDIFFEQPACK} 0.1.0, \textsc{CUDA} 11.2, and \textsc{NVIDIA} Driver 460.91.03, and i7 CPU, and \textsc{NVIDIA Quadro RTX 8000}.

\begin{table*}[t]
\footnotesize
\renewcommand{\arraystretch}{0.86}
\setlength{\tabcolsep}{2pt}
\caption{Best hyperparameters and the number of trainable parameters. The weight decay is the $L^2$ regularization coefficient term $w$ of ~\eqref{eq:loss}. The number of final layers is $M$ of ~\eqref{eq:final}. Considering the original price range of each dataset, we set the initial coefficient value of the price reflection module, $\bm{c}$ of ~\eqref{eq:price}.}
\label{tab:param}
\begin{tabular}{c|c|l|c|c|c|c|c|c}
\toprule
\multirow{2}{*}{}       & \multicolumn{2}{c|}{Model} & Batch size & Learning rate & Weight decay & {\#Final layers} & Initial coefficient & {\#Params} \\ \midrule
\multirow{8}{*}{\rotatebox[origin=c]{90}{San Francisco}} & \multirow{6}{*}{Substitute Eq.~\eqref{eq:short} with} 
                                 & RNN   & 64 & \num{1e-4} & \num{1e-5} & 10 & 0.03 & 7,911,217\\
                               & & LSTM  & 64 & \num{1e-4} & \num{1e-5} & 10 & 0.03 & 10,404,217 \\
                               & & GRU   & 64 & \num{1e-4} & \num{1e-5} & 10 & 0.03 & 9,573,217 \\
                               & & STGCN & 64 & \num{1e-3} & \num{1e-5} & 10 & 0.03 & 7,700,854\\
                               & & DCRNN & 64 & \num{1e-3} & \num{1e-5} & 10 & 0.03 & 7,753,914\\
                               & & AGCRN & 64 & \num{1e-3} & \num{1e-5} & 10 & 0.03 & 13,594,000\\ \cmidrule{2-9} 
                               &\multicolumn{2}{l|}{\textbf{Proposed (full model)}} & 64 & \num{1e-4} & \num{1e-5} & 15 & 0.03 & 7,001,059 \\ 
                               &\multicolumn{2}{l|}{\textbf{Proposed (w/o final module)}} & 64 & \num{1e-4} & \num{1e-5} & 15 & 0.03 & 8,131,549 \\ \midrule
\multirow{8}{*}{\rotatebox[origin=c]{90}{Seattle}} & \multirow{6}{*}{Substitute Eq.~\eqref{eq:short} with} 
                                  & RNN  & 64 & \num{1e-3} & \num{1e-5} & 10 & 0.5 & 3,544,657 \\
                               & & LSTM  & 64 & \num{1e-3} & \num{1e-5} & 10 & 0.5 & 5,947,657 \\
                               & & GRU   & 64 & \num{1e-3} & \num{1e-5} & 10 & 0.5 & 5,146,657 \\
                               & & STGCN & 64 & \num{1e-3} & \num{1e-5} & 10 & 0.5 & 2,977,894\\
                               & & DCRNN & 64 & \num{1e-3} & \num{1e-5} & 10 & 0.5 & 3,031,074 \\
                               & & AGCRN & 64 & \num{1e-3} & \num{1e-5} & 10 & 0.5 & 5,322,406\\ \cmidrule{2-9} 
                               &\multicolumn{2}{l|}{\textbf{Proposed (full model)}} & 64 & \num{5e-4} & \num{5e-5} & 10 & 0.5 & 2,985,619 \\ 
                               &\multicolumn{2}{l|}{\textbf{Proposed (w/o final module)}} & 64 & \num{5e-4} & \num{5e-5} & 10 & 0.5 & 2,694,559 \\  
\bottomrule
\end{tabular}
\end{table*}

\subsection{Experimental Environments}
\paragraph{Dataset} The San Francisco Municipal Transportation Agency (SFMTA) website \footnote{\url{https://www.sfmta.com/getting-around/drive-park/demand-responsive-pricing/sfpark-evaluation}} provides data collected during the SFpark pilot project. On-street occupancy rate data contain per-block hourly occupancy rate and meter prices for seven parking districts.


The Seattle Department of Transportation (SDOT) sets on-street parking rates based on data to obtain a target occupancy rate of 70\% to 85\%. The SDOT website \footnote{\url{https://www.seattle.gov/transportation/projects-and-programs/programs/parking-program/performance-based-parking-pricing-program}} provides a dataset of on-street occupancy rates from 2010. In contrast to San Francisco, Seattle has fixed annual pricing that is updated based on the previous year's data. Prices vary depending on the time of day; 8 a.m. to 11 a.m., 11 a.m. to 5 p.m., and 5 p.m. to 8 p.m. 


Since we optimize the parking price, we define parking occupancy as metered parking occupancy and only predict and optimize when the parking blocks are metered. For instance, as shown in Fig. ~\ref{meter_oper}, meters are only operational from 9 a.m. to 5 p.m in San Francisco. Many parking blocks in Seattle also are not metered after 5 p.m. In San Francisco, parking is free on all weekends, and in Seattle on Sundays. Therefore, only data from 9 a.m. to 5 p.m. of San Francisco and data from 8 a.m. to 5 p.m. of Seattle on weekdays are used in our experiments.

We use data from August 11, 2011, to July 31, 2013 for San Francisco. Finally, $N=158$ blocks from 63 streets and 7 districts with no missing data in the aforementioned period are selected for prediction/optimization. For Seattle, we extract the most recent data from July 19, 2019 to March 23, 2020 and predict/optimize $N=98$ parking blocks in 9 parking areas.





Historical data from various periods are derived via feature engineering. In the case of short-term history, the occupancy rate in the past $K$ hours is given as a feature. Different $K$ settings lead to different training/testing data as shown in Table~\ref{dataset}. For example, San Francisco dataset consists of 407,008 (resp. 356,132) samples, when $K=1$ (resp. $K=2$). The ratio between a training set and a test set is 7:3. We only consider $K=\{1,2,3\}$ for short-term history length in our experiments because longer $K$ settings lead to a drastic reduction in the number of training samples (as daily occupancy rates are only available from 8 a.m./9 a.m. to 5 p.m.). However, long-term history contains much detailed information.






For the long-term historical features, we obtained historical occupancy records for various periods. The average occupancy rates for the last week, last two weeks, last month, and past two months are derived. Moreover, we separated the time zones into 9 a.m. to 11 a.m., 12 p.m. to 2 p.m., and 3 p.m. to 5 p.m. Adopting this strategy, we ended up with 12 long-term history features. 

\paragraph{Baselines} We compare our model with the following baseline prediction models:
\begin{compactenum}
    \item RNN, LSTM~\cite{sepp1997long} and GRU~\cite{chung2014empirical} are popular time-series processing models. We create two different ways of how to utilize them for experiments: i) These models are used instead of our short-term history layer and other implementations are the same. ii) We feed only the short-term historical information to them and predict (without any other layers) since they are designed to consider only one type of sequence.
    \item DCRNN~\citep{li2018dcrnn_traffic}, AGCRN~\citep{NEURIPS2020_ce1aad92}, and STGCN~\citep{bing2018stgcn} are popular baselines in the field of spatiotemporal machine learning. DCRNN combines graph convolution with recurrent neural networks in an encoder-decoder manner. AGCRN learns an adaptive adjacency matrix from data. STGCN combines graph convolution with gated temporal convolution. We also define two different versions for them in the same way above.
\end{compactenum}

\begin{table*}[t]
\footnotesize
\setlength{\tabcolsep}{2pt}
\renewcommand{\arraystretch}{0.86}
\caption{The results of parking occupancy prediction (mean $\pm$ std.dev.)}
\label{prediction-results}
\begin{tabular}{c|c|c|l|cc|cc|cc}
\toprule
\multirow{2}{*}{}       & \multicolumn{3}{c|}{\multirow{3}{*}{Model}} & \multicolumn{2}{c|}{$K=1$}            & \multicolumn{2}{c|}{$K=2$}                                   & \multicolumn{2}{c}{$K=3$}                                   \\  \cmidrule{5-10} 
                               &  \multicolumn{3}{l|}{}     & MSE   & $R^2$ & \multicolumn{1}{c}{MSE}   & \multicolumn{1}{c|}{$R^2$}    & \multicolumn{1}{c}{MSE}   & \multicolumn{1}{c}{$R^2$}\\ \midrule
\multirow{16}{*}{\rotatebox[origin=c]{90}{San Francisco}} &\multirow{8}{*}{\textbf{Ours}} & \multirow{6}{*}{Substitute Eq.~\eqref{eq:short} with} 
                                  & RNN   & 0.01374 $\pm$ 0.00027 & 0.60727 $\pm$ 0.00763 & 0.01296 $\pm$ 0.00223 & 0.62080 $\pm$ 0.00223 & 0.01373 $\pm$ 0.00016 & 0.59760 $\pm$ 0.00457 \\
                               && & LSTM  & 0.01701 $\pm$ 0.00027 & 0.51369 $\pm$ 0.01161  & 0.01443 $\pm$ 0.00321 & 0.57783 $\pm$ 0.00321 & 0.01517 $\pm$ 0.00009 & 0.55557 $\pm$ 0.00261 \\
                               && & GRU   & 0.01505 $\pm$ 0.00039 & 0.56983 $\pm$ 0.01113 & 0.01395 $\pm$ 0.00334 & 0.59187 $\pm$ 0.00334 & 0.01446 $\pm$ 0.00014 & 0.57618 $\pm$ 0.00406 \\
                               && & STGCN & 0.01040 $\pm$ 0.00050 & 0.70287 $\pm$ 0.01419 & 0.01027 $\pm$ 0.00045 & 0.69969 $\pm$ 0.01319 & 0.01037 $\pm$ 0.00049 & 0.69619 $\pm$ 0.01433 \\
                               && & DCRNN & 0.01022 $\pm$ 0.00030 & 0.70796 $\pm$ 0.00851 & 0.01012 $\pm$ 0.00041 & 0.70404 $\pm$ 0.01199 & 0.01020 $\pm$ 0.00053 & 0.70113 $\pm$ 0.01534 \\
                               && & AGCRN & 0.01021 $\pm$ 0.00012 & 0.70821 $\pm$ 0.00329 & 0.01001 $\pm$ 0.00014 & 0.70727 $\pm$ 0.00417 & 0.01047 $\pm$ 0.00032 & 0.69326 $\pm$ 0.00938 \\ \cmidrule{3-10} 
                               &&\multicolumn{2}{l|}{\textbf{Proposed (full model)}} & \textbf{0.00980 $\pm$ 0.00002} & \textbf{0.71985 $\pm$ 0.00046} & \textbf{0.00975 $\pm$ 0.00001} & \textbf{0.71473 $\pm$ 0.00034} & \textbf{0.00999 $\pm$ 0.00003}      & \textbf{0.70726 $\pm$ 0.00092}            \\ 
                               &&\multicolumn{2}{l|}{\textbf{Proposed (w/o final module)}} & 0.01005 $\pm$ 0.00003 & 0.71278 $\pm$ 0.00084 & 0.00994 $\pm$ 0.00004 & 0.70915 $\pm$ 0.00109 & 0.01009 $\pm$ 0.00003 & 0.70435 $\pm$ 0.00090 \\ \cmidrule{2-10}
                               &\multirow{7}{*}{\begin{tabular}[c]{@{}c@{}}Existing\\Method\\(only\\short-term)\end{tabular}}
                               & \multicolumn{2}{c|}{RNN}   & 0.01295 $\pm$ 0.00004 & 0.62994 $\pm$ 0.00120 & 0.01258 $\pm$ 0.00003 & 0.63189 $\pm$ 0.00102 & 0.01321 $\pm$ 0.01378 & 0.61225 $\pm$ 0.00145 \\
                               && \multicolumn{2}{c|}{LSTM} & 0.01564 $\pm$ 0.00020 & 0.55307 $\pm$ 0.00562 & 0.01395 $\pm$ 0.00005 & 0.59198 $\pm$ 0.00137 & 0.01476 $\pm$ 0.01378 & 0.56762 $\pm$ 0.00296 \\
                               && \multicolumn{2}{c|}{GRU}  & 0.01374 $\pm$ 0.00006 & 0.60724 $\pm$ 0.00174 & 0.01280 $\pm$ 0.00006 & 0.62547 $\pm$ 0.00167 & 0.01319 $\pm$ 0.01378 & 0.61344 $\pm$ 0.00240 \\
                               && \multicolumn{2}{c|}{STGCN}& 0.01119 $\pm$ 0.00054 & 0.68005 $\pm$ 0.01546 & 0.01100 $\pm$ 0.00047 & 0.67825 $\pm$ 0.01378 & 0.01112 $\pm$ 0.00077 & 0.67407 $\pm$ 0.02243 \\
                               && \multicolumn{2}{c|}{DCRNN}& 0.01125 $\pm$ 0.00004 & 0.67850 $\pm$ 0.00127 & 0.01077 $\pm$ 0.00002 & 0.68493 $\pm$ 0.00046 & 0.01072 $\pm$ 0.00005 & 0.68578 $\pm$ 0.00136 \\
                               && \multicolumn{2}{c|}{AGCRN}& 0.01092 $\pm$ 0.00010 & 0.68796 $\pm$ 0.00282 & 0.01059 $\pm$ 0.00012 & 0.69012 $\pm$ 0.00355 & 0.01095 $\pm$ 0.00066 & 0.67913 $\pm$ 0.01948 \\  
                               && \multicolumn{2}{c|}{NODE} & \textbf{0.01058 $\pm$ 0.00005} & \textbf{0.69753 $\pm$ 0.00146} & \textbf{0.01055 $\pm$ 0.00008} & \textbf{0.69123 $\pm$ 0.00238} & \textbf{0.01063 $\pm$ 0.00006} & \textbf{0.68840 $\pm$ 0.00170} \\ \midrule
\multirow{16}{*}{\rotatebox[origin=c]{90}{Seattle}} &\multirow{8}{*}{\textbf{Ours}} & \multirow{6}{*}{Substitute Eq.~\eqref{eq:short} with} 
                                & RNN       & 0.02392 $\pm$ 0.00015 & 0.62621 $\pm$ 0.00225 & 0.02721 $\pm$ 0.00036 & 0.57511 $\pm$ 0.00567 & 0.02799 $\pm$ 0.00053 & 0.56240 $\pm$ 0.00828 \\
                              &&& LSTM      & 0.02564 $\pm$ 0.00034 & 0.59918 $\pm$ 0.00529 & 0.02797 $\pm$ 0.00032 & 0.56323 $\pm$ 0.00500 & 0.03128 $\pm$ 0.00085 & 0.51106 $\pm$ 0.01323 \\
                              &&& GRU       & 0.02466 $\pm$ 0.00010 & 0.61449 $\pm$ 0.00151 & 0.02660 $\pm$ 0.00058 & 0.58451 $\pm$ 0.00902 & 0.02917 $\pm$ 0.00149 & 0.54399 $\pm$ 0.01746 \\
                              &&& STGCN     & 0.02125 $\pm$ 0.00029 & 0.66792 $\pm$ 0.00451 & 0.02154 $\pm$ 0.00103 & 0.66360 $\pm$ 0.01611 & 0.02183 $\pm$ 0.00112 & 0.65870 $\pm$ 0.01746 \\
                              &&& DCRNN     & 0.02128 $\pm$ 0.00023 & 0.66740 $\pm$ 0.00352 & 0.02171 $\pm$ 0.00056 & 0.66094 $\pm$ 0.00873 & 0.02203 $\pm$ 0.00088 & 0.65556 $\pm$ 0.01376 \\
                              &&& AGCRN     & 0.02165 $\pm$ 0.00009 & 0.66162 $\pm$ 0.00140 & 0.02170 $\pm$ 0.00038 & 0.66137 $\pm$ 0.00643 & 0.02359 $\pm$ 0.00038 & 0.63345 $\pm$ 0.00610 \\ \cmidrule{3-10} 
                               &&\multicolumn{2}{l|}{\textbf{Proposed (full model)}} & \textbf{0.02098 $\pm$ 0.00006} & \textbf{0.67204 $\pm$ 0.00088} & \textbf{0.02126 $\pm$ 0.00008} & \textbf{0.66803 $\pm$ 0.00122} & \textbf{0.02153 $\pm$ 0.0000} & \textbf{0.66339 $\pm$ 0.00029} \\ 
                               &&\multicolumn{2}{l|}{\textbf{Proposed (w/o final module)}} & 0.02266 $\pm$ 0.00009 & 0.64580 $\pm$ 0.00140 & 0.02331 $\pm$ 0.00017 & 0.63599 $\pm$ 0.00263 & 0.02386 $\pm$ 0.00021 & 0.62708 $\pm$ 0.00327 \\ \cmidrule{2-10}
                              &\multirow{7}{*}{\begin{tabular}[c]{@{}c@{}}Existing\\Method\\(only\\short-term)\end{tabular}}
                              & \multicolumn{2}{c|}{RNN}   & 0.02428 $\pm$ 0.00002 & 0.62045 $\pm$ 0.00035 & 0.02674 $\pm$ 0.00010 & 0.58230 $\pm$ 0.00162 & 0.02712 $\pm$ 0.00010 & 0.57598 $\pm$ 0.00155 \\
                              && \multicolumn{2}{c|}{LSTM} & 0.02490 $\pm$ 0.00008 & 0.61076 $\pm$ 0.00118 & 0.02529 $\pm$ 0.00013 & 0.60497 $\pm$ 0.00202 & 0.02720 $\pm$ 0.00004 & 0.57485 $\pm$ 0.00063 \\
                              && \multicolumn{2}{c|}{GRU}  & 0.02427 $\pm$ 0.00006 & 0.62067 $\pm$ 0.00088 & 0.02466 $\pm$ 0.00004 & 0.61491 $\pm$ 0.00064 & 0.02548 $\pm$ 0.00010 & 0.60162 $\pm$ 0.00156 \\
                              && \multicolumn{2}{c|}{STGCN}& 0.02148 $\pm$ 0.00037 & 0.66423 $\pm$ 0.00574 & 0.02167 $\pm$ 0.00007 & 0.66158 $\pm$ 0.00111 & 0.02179 $\pm$ 0.00008 & 0.65931 $\pm$ 0.00131 \\
                              && \multicolumn{2}{c|}{DCRNN}& 0.02416 $\pm$ 0.00022 & 0.62232 $\pm$ 0.00338 & 0.02323 $\pm$ 0.00047 & 0.63721 $\pm$ 0.00735 & 0.02250 $\pm$ 0.00036 & 0.64833 $\pm$ 0.00558 \\
                              && \multicolumn{2}{c|}{AGCRN}& 0.02158 $\pm$ 0.00010 & 0.66263 $\pm$ 0.00158 & 0.02485 $\pm$ 0.00217 & 0.61193 $\pm$ 0.03388 & \textbf{0.02168 $\pm$ 0.00021} & \textbf{0.66108 $\pm$ 0.00321} \\ 
                              && \multicolumn{2}{c|}{NODE} & \textbf{0.02138 $\pm$ 0.00006} & \textbf{0.66596 $\pm$ 0.00094} &\textbf{ 0.02163 $\pm$ 0.00005} & \textbf{0.66211 $\pm$ 0.00078} & 0.02182 $\pm$ 0.00005 & 0.65894 $\pm$ 0.00080 \\ 
\bottomrule                               
\end{tabular}
\end{table*}

\paragraph{Hyperparameters} We test the following hyperparameters for each model --- we trained for 1,000 epochs with a batch size of 64 for all models and other detailed settings are in Table~\ref{tab:param}:
\begin{compactenum}
    \item For RNN, LSTM, and GRU, we use one layer has the hidden size of 500.
    \item For STGCN, we set the channels of three layers (two temporal gated convolution layers and one spatial graph convolution layer) in ST-Conv block to 64, and the Chebyshev polynomials to 1. To construct a parking block graph, we compute the pairwise network distances between blocks and build an adjacency matrix using thresholded Gaussian kernel. 
    \item For DCRNN, we set the number of GRU hidden units to 64 and the number of recurrent layers to 2. We use the weighted adjacency matrix constructed in the same way as STGCN.
    \item For AGCRN, we set the hidden unit to 64 for all the AGCRN cells. The size of node embeddings is set to 2 and 5 for San Francisco and Seattle, respectively.
    \item For ours, NODE, we use MALI~\cite{zhuang2021mali}, the training algorithm for NODEs and the ODE solver error tolerance is set to 1e-5.
\end{compactenum}

\subsection{Experimental Results}

Experimental results are summarized in Table~\ref{prediction-results} --- we report the mean and std. dev. performance with five different random seeds. In general, RNN-based models perform poorly compared to other spatiotemporal models, such as AGCRN and STGCN. 
Moreover, when spatiotemporal models are plugged into our model in the place of Eq.~\eqref{eq:short}, they mostly perform better than when they are used solely with the short-term information. Other modules remain the same when we substitute Eq.~\eqref{eq:short} with other baselines, thus the long-term features and price are also used, which is reasonable. In all cases, the best outcomes are achieved by our full model.

\paragraph{Ablation Study}
As an ablation study, we remove the final prediction module and let $\bm{z}_{adjust}$ be the final prediction. Since $\bm{z}_{adjust}$ is adjusted with the price information from the initial prediction, we can also use it for price optimization (or dynamic pricing). However, the proposed prediction model without the final prediction module does not perform as well as our full model as shown in Table~\ref{prediction-results}.

\section{Experiments on Optimization}
We describe our optimization results. The prediction model is fixed during this process and we optimize an actionable input, i.e., price, to achieve target occupancy rates. Since our proposed full model produces the lowest prediction errors, we use it as an oracle. In this step, it is crucial to use the best predictive model as an oracle (although some sub-optimal models provide a lower complexity)~\cite{An:2016:MFM:2939672.2939726,An:2017:DFA:3055535.3041217,AAAI18-Chen,li2021largescale,doi:10.1137/1.9781611976700.80,parkape}. If not, the optimized solution (toward the sub-optimal model) does not yield expected outcomes when being applied to real-world environments.

\subsection{Experimental Environments}

\begin{table}[t]
\small
\setlength{\tabcolsep}{2pt}
\renewcommand{\arraystretch}{0.8}
\caption{The optimization performance (the ratio of failed test cases where the optimized occupancy rate exceeds the threshold $\tau$) and the average runtime of San Francisco}
\label{opt-performance}
\begin{tabular}{l|ccc|c}
\toprule
                            & \multicolumn{3}{c|}{\bfseries Optimization Performance} & \multirow{2}{*}{\begin{tabular}[c]{@{}c@{}}\bfseries Runtime\\ \bfseries (seconds)\end{tabular}} \\
                            & $\bm{\tau = 70\%}$         & $\bm{\tau = 75\%}$        & $\bm{\tau = 80\%}$         &      \\ \midrule
Observed in Data            & 0.5475                     & 0.4440                     & 0.3450                     & N/A                   \\ 
Greedy                      & 0.4685                     & 0.2045                     & 0.0553                     & 446.6343           \\
Gradient-based              & 0.3606                     & 0.1609                     & 0.0488                     & 0.5471               \\ \midrule
\bfseries One-shot (Ours)   & \bfseries 0.1938          & \bfseries 0.0065            & \bfseries 0.0033           & \bfseries 0.0000007       \\
\bottomrule
\end{tabular}
\end{table}

\begin{table}[t]
\small
\setlength{\tabcolsep}{2pt}
\renewcommand{\arraystretch}{0.8}
\caption{The optimization performance and the average runtime of Seattle}
\label{opt-performance-seattle}
\begin{tabular}{l|ccc|c}
\toprule
                            & \multicolumn{3}{c|}{\bfseries Optimization Performance} & \multirow{2}{*}{\begin{tabular}[c]{@{}c@{}}\bfseries Runtime\\ \bfseries (seconds)\end{tabular}} \\
                            & $\bm{\tau = 70\%}$        & $\bm{\tau = 75\%}$        & $\bm{\tau = 80\%}$       &      \\ \midrule
Observed in Data            & 0.1307                    & 0.1003                    & 0.0756                   & N/A                   \\ 
Greedy                      & 0.1533                    & 0.0917                    & 0.0495                   & 0.5438           \\
Gradient-based              & 0.1739                    & 0.1171                    & 0.0724                   & 0.0042               \\ \midrule
\bfseries One-shot (Ours)   & \bfseries 0.0997          & \bfseries 0.0684          & \bfseries 0.0440          & \bfseries 0.0000007       \\
\bottomrule
\end{tabular}
\end{table}

\paragraph{Baselines} We compare our method with the following baseline methods (whose design concepts are similar to what used in~\cite{An:2016:MFM:2939672.2939726,An:2017:DFA:3055535.3041217,li2021largescale,doi:10.1137/1.9781611976700.80})  --- the target occupancy rate $\bm{y}^*$ is set to 70\% and the minimum price $p_{min}$ is set to \$0.25 and \$0.5 for San Francisco and Seattle, respectively; the maximum price $p_{max}$ is set to \$34.5 for San Francisco and \$3.0 for Seattle. Given the short-term, long-term occupancy rates, and the pre-trained prediction model's parameters $\bm{\theta}_{\xi}$, we optimize the price $\bm{p}^*$ given the target occupancy rate $\bm{y}^*$:
\begin{compactenum}
    \item We use the following greedy black-box search strategy: We set the initial price vector $\bm{p}^*$ to $p_{min}$ for all parking blocks. For each iteration, we increase each block's parking price by \$0.25 (given that its price has not already reached $p_{max}$) and find the parking block which decreases the error term $\|\bm{y}^* - \hat{\bm{y}}\|^2_2$ the most. We then increase the price of this block by \$0.25. The aforementioned strategy is repeated until the current iteration does not decrease the error term.
    \item We test the following gradient-based white-box strategy: The gradient $\frac{\partial \|\bm{y}^* - \hat{\bm{y}}\|^2_2}{\partial \bm{p}^*}$ flows to $\bm{p}^*$ with the Adam optimizer~\cite{kingma2014adam} up to 1,000 iterations until the error term $\|\bm{y}^* - \hat{\bm{y}}\|^2_2$ stabilizes --- $\bm{p}^*$ is initialized to \$0.25 for all parking blocks. Note that in this scheme, $\bm{p}^*$ is optimized to minimize the error term, with $p_{min} \leq \bm{p}^* \leq p_{max}$. This method can find the optimal solution if the error term is a convex function, which is not the case in our setting, so it returns a sub-optimal solution.
\end{compactenum}

The worst-case complexity of the greedy strategy is $\mathcal{O}(N^2\frac{p_{max} - p_{min}}{0.25})$ whereas our method has a complexity of $\mathcal{O}(1)$. Depending on how quickly the solution converges, the gradient method's performance varies substantially.

\paragraph{Research Questions} We answer the following research questions from our experiments:
\begin{compactenum}
    \item []\hspace{-2em}(RQ1) What is the runtime of each method?
    \item []\hspace{-2em}(RQ2) For how many test cases does each method successfully find (or fail to find) $\bm{p}^*$?
\end{compactenum}

\subsection{Experimental Results}

\paragraph{RQ1} In Tables~\ref{opt-performance} and~\ref{opt-performance-seattle}, we compare the optimization results. In the case of San Francisco, our method is several orders of magnitude faster than the greedy black-box method and the gradient-based white box method. The greedy method, in particular, showed significant limitations in terms of runtime, with one test case lasting 447 seconds on average and taking up to 4,653 iterations. Considering that we optimized 183,280 cases in the test period, our one-shot method showed remarkable runtime. Seattle's price range is not as broad as San Francisco's, i.e., \$0.5 to \$3.0. As a result, the optimization of Seattle showed faster runtime in Table~\ref{opt-performance-seattle}.

\paragraph{RQ2} We also show the percentage of the failure cases where the optimized occupancy rate is larger than the threshold $\tau = \{70\%, 75\%, 80\%\}$. These errors occur due to imperfections in the price optimization process. Our method greatly outperforms other methods by showing a much smaller failure ratio in comparison to others. Theoretically, our method should show zero error if we can exactly solve the reverse-mode integral, but due to the practical limitation of ODE solvers, it produces a small amount of error.  However, 80.6\% to 99.7\% of the 183,280 cases in San Francisco, and 91\% to 95.6\% of the 31,360 cases in Seattle are still below the optimal occupancy rate $\tau = \{70\%, 75\%, 80\%\}$, which are acceptable.

\begin{figure}[t]
    \centering
    \includegraphics[width=0.9\columnwidth]{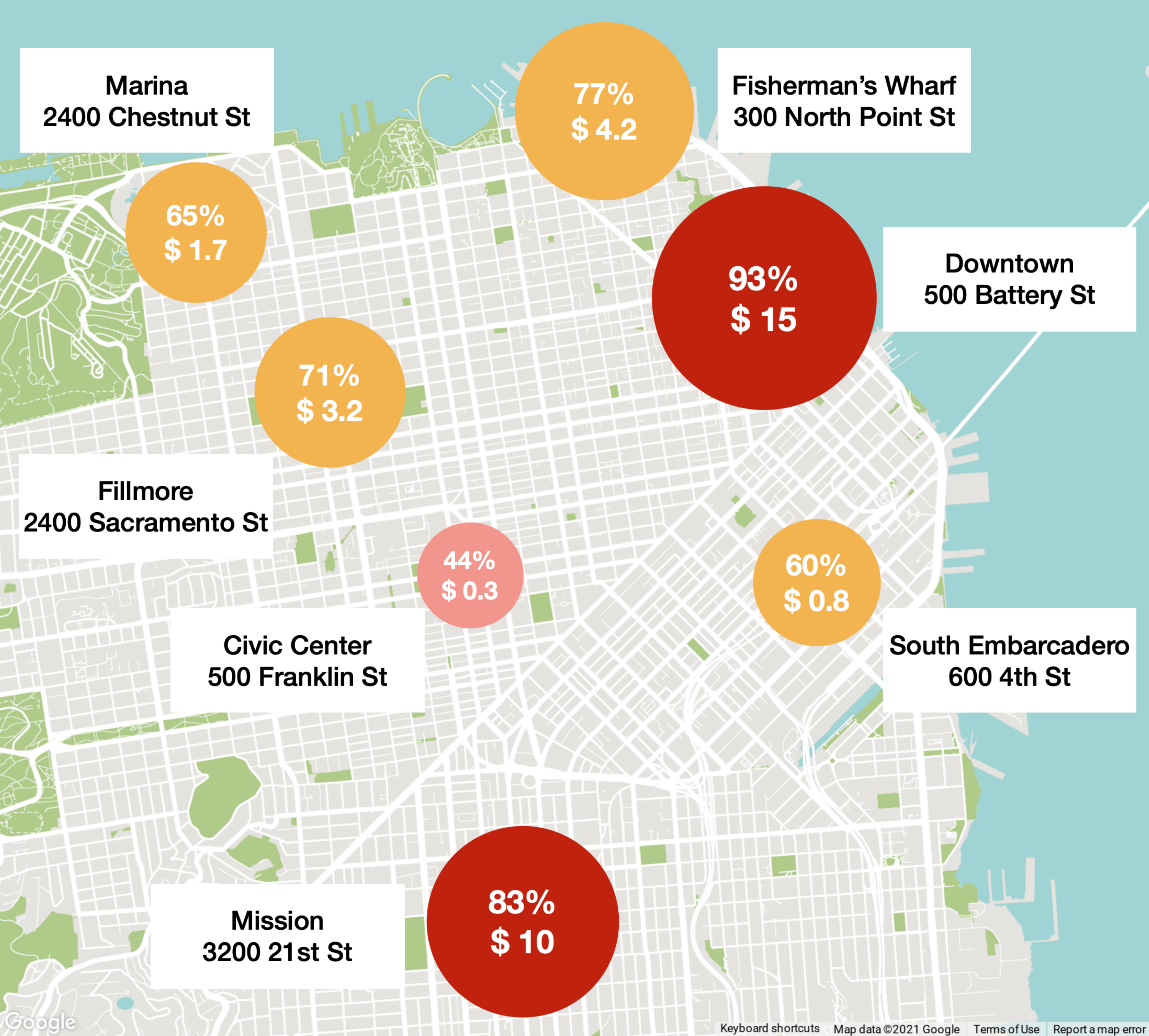}
    \caption{San Francisco's mean observed occupancy rate and the mean optimized parking price at 12:00 p.m. in the test period for seven blocks in seven different districts}
    \label{opt_map_viz}
\end{figure}

\section{Case Studies}

We introduce selected examples of prediction and price optimization outcomes with visualization. Figure~\ref{opt_map_viz} shows the mean \emph{observed} (or ground-truth in the dataset) occupancy rate and \emph{optimized} parking price of San Francisco during the test period. As shown, the optimized parking price and the observed parking occupancy rate are highly correlated. The correlation between the mean hourly observed occupancy rate and the mean optimized price on each block is 0.724 in our San Francisco's test set.

The interesting point is that the mean optimized price jumped when the observed occupancy exceeded the ideal rate of 80\%. The parking block in downtown, 500 Battery St's occupancy rate was almost always over 90\% at noon, with a mean rate of 93\%. The mean optimized price at noon is \$15, a fairly high price. Since the parking block is almost always over-occupied, our framework accordingly recommends a reasonably high price to reduce the occupancy rate. The average parking price observed in the dataset was \$5.9, which seems to be insufficiently high to achieve the target occupancy rate.

Our strategy also shows adaptability to sudden changes in demand. On July 4, 2013, the aforementioned block's parking occupancy rate was observed 1\% at noon. For this day, our method dynamically decreased the price to \$0.25, the minimum price. However, the observed price was \$6. Similarly, the block in the civic center, 500 Franklin St., showed a low occupancy rate on average. However, on April 12, the occupancy rate at noon was higher than usual at 81\%. Our optimized price was \$1.72, while the observed price was \$0.75--even lower than the mean price of \$0.86. These cases demonstrate the robustness of our proactive optimization.

\begin{figure}[t]
    \setlength{\abovecaptionskip}{10pt}
    \setlength{\belowcaptionskip}{-5pt}
    \centering
    \subfigure[The block between E James St and E Cherry St, 12th Avenue]{\includegraphics[width=0.8\columnwidth]{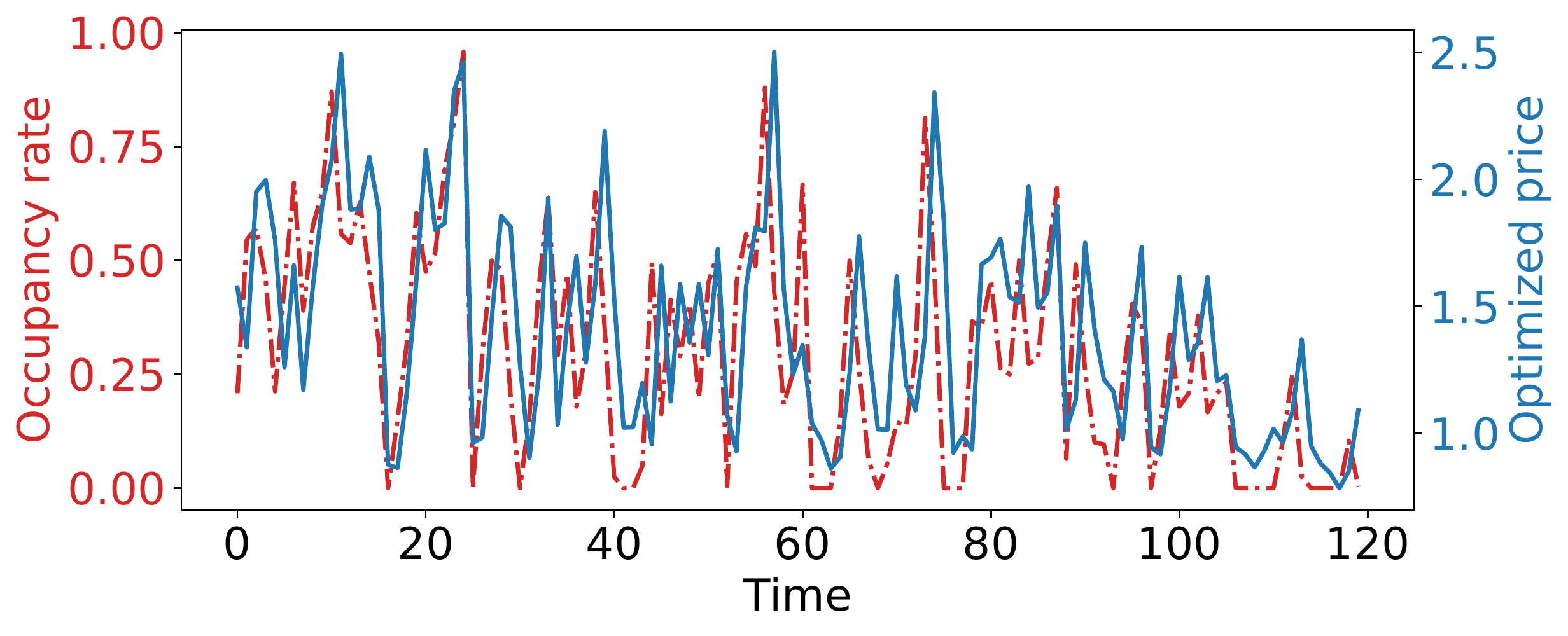}}
    \subfigure[The block between E Marison St and E Spring St, 12th Avenue]{\includegraphics[width=0.8\columnwidth]{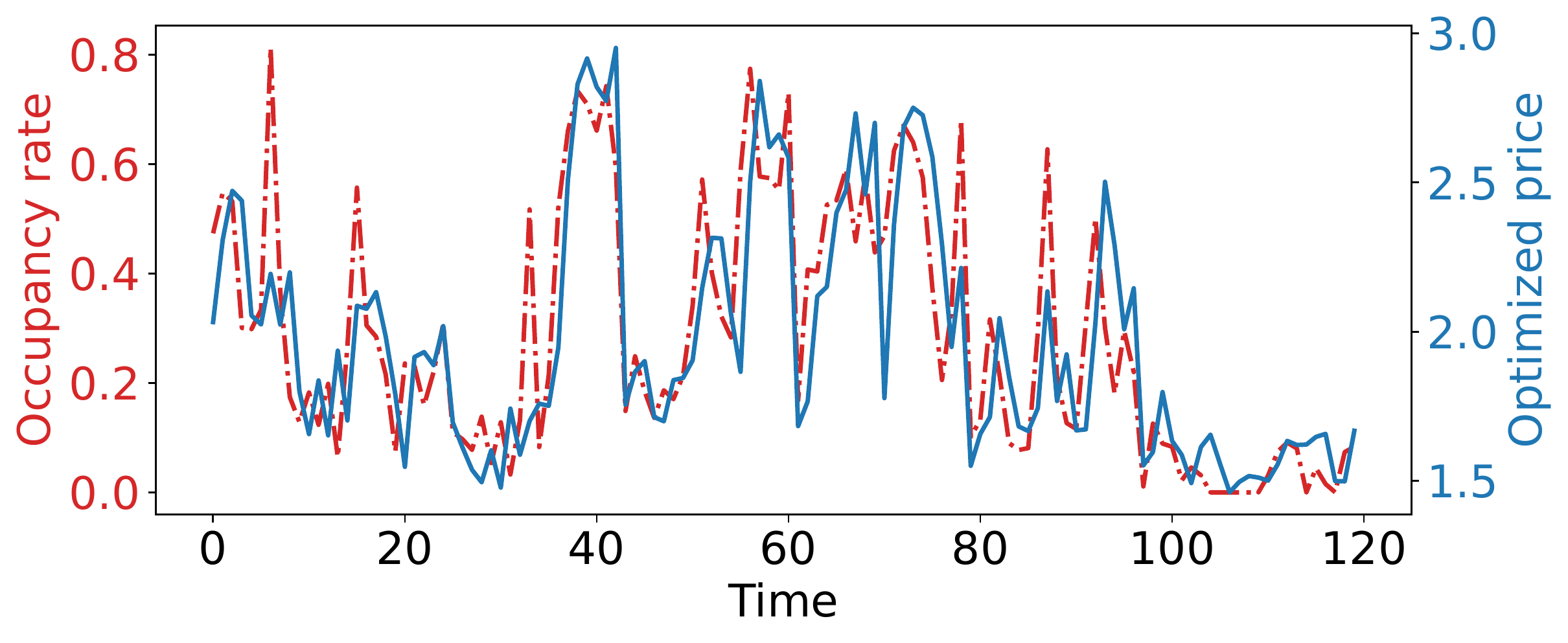}}
    \caption{The correlation visualization between the occupancy rate and the optimized price. These figures include Seattle's test data from January 31st.}
    \label{opt_corr_viz}
\end{figure}

Fig.~\ref{opt_corr_viz} shows the effectiveness of our dynamic pricing with some selected examples in two parking blocks in Seattle. As shown, their occupancy rates by the oracle in red are well controlled by our dynamic pricing in blue. The observed ground-truth price of these blocks is set to the basic price, \$0.5, and adjusted to \$1 at 5 p.m.

\begin{figure}[t]
    \setlength{\abovecaptionskip}{10pt}
    \setlength{\belowcaptionskip}{-4pt}
    \centering
    \subfigure[East Green Lake Dr N between NE 72nd St and NE Maple Leaf Pl]{\includegraphics[width=0.23\textwidth]{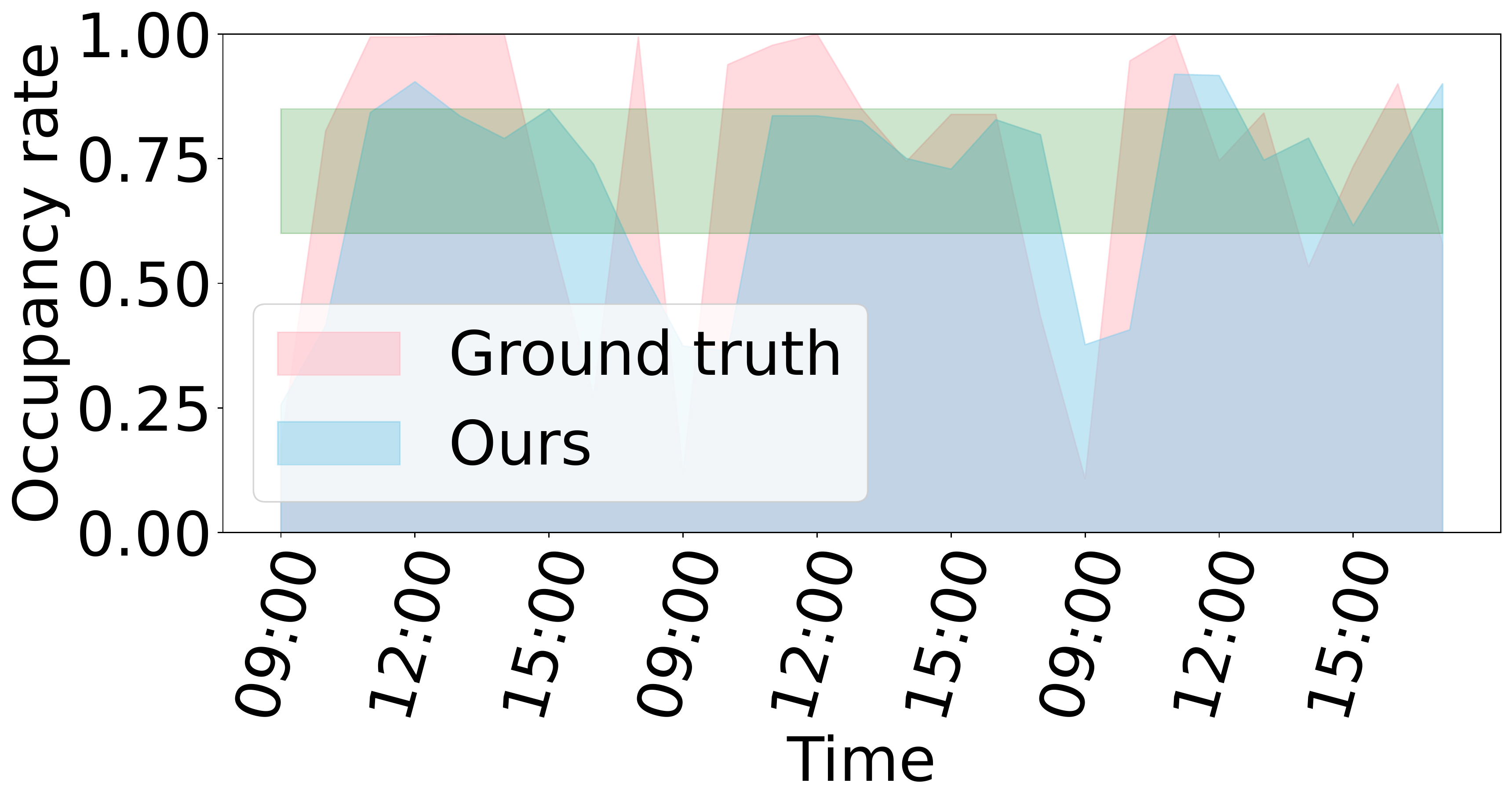}} \hfill
    \subfigure[East Green Lake Dr N between NE Ravenna EB Blvd and NE 72nd St]{\includegraphics[width=0.23\textwidth]{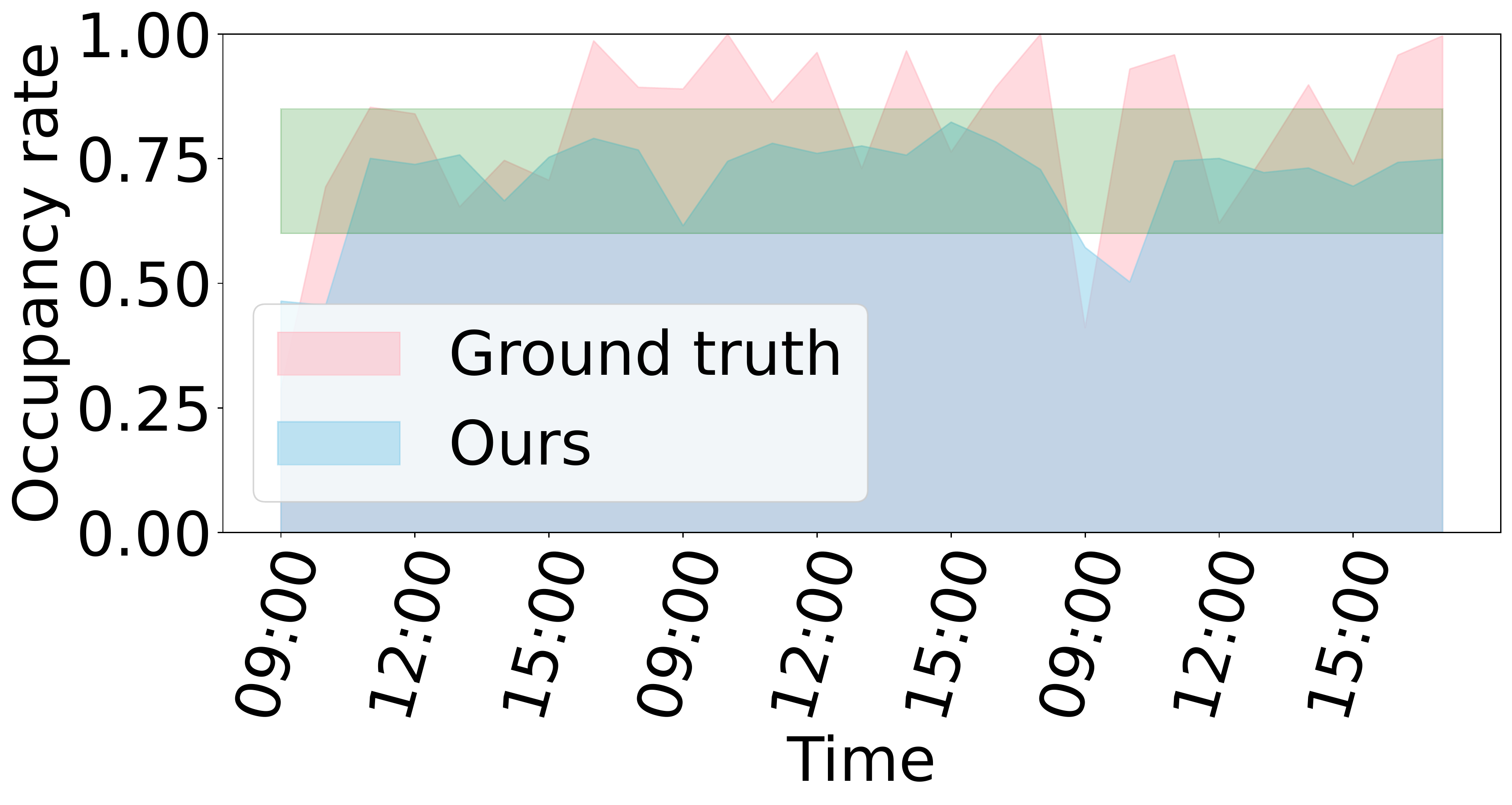}} \\
    \subfigure[East Green Lake Way N between 4th Ave NEand NE Ravenna SR Blvd]{\includegraphics[width=0.23\textwidth]{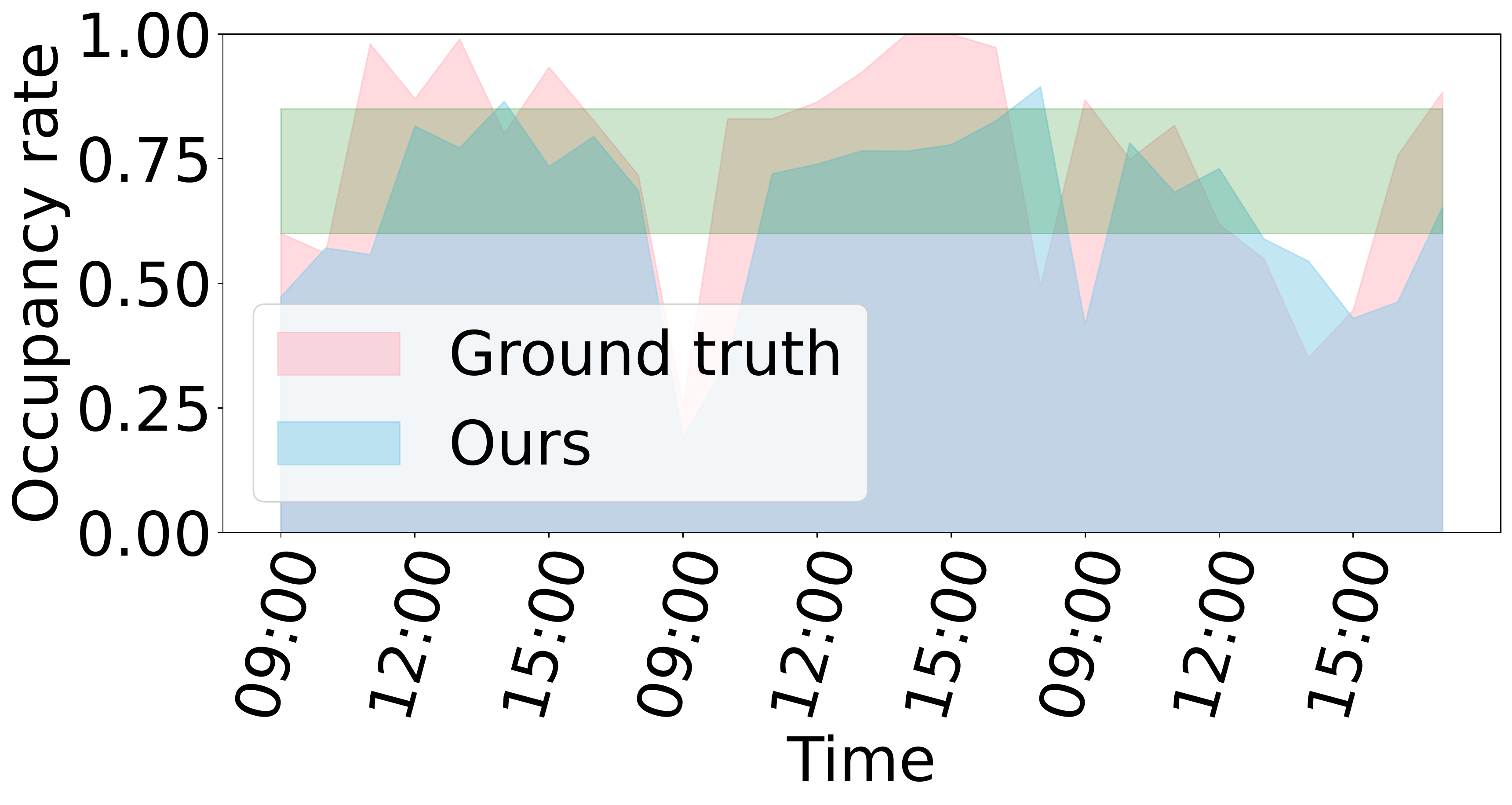}} \hfill
    \subfigure[NE 64th St between 9th Ave NE and Roosebelt Way NE]{\includegraphics[width=0.23\textwidth]{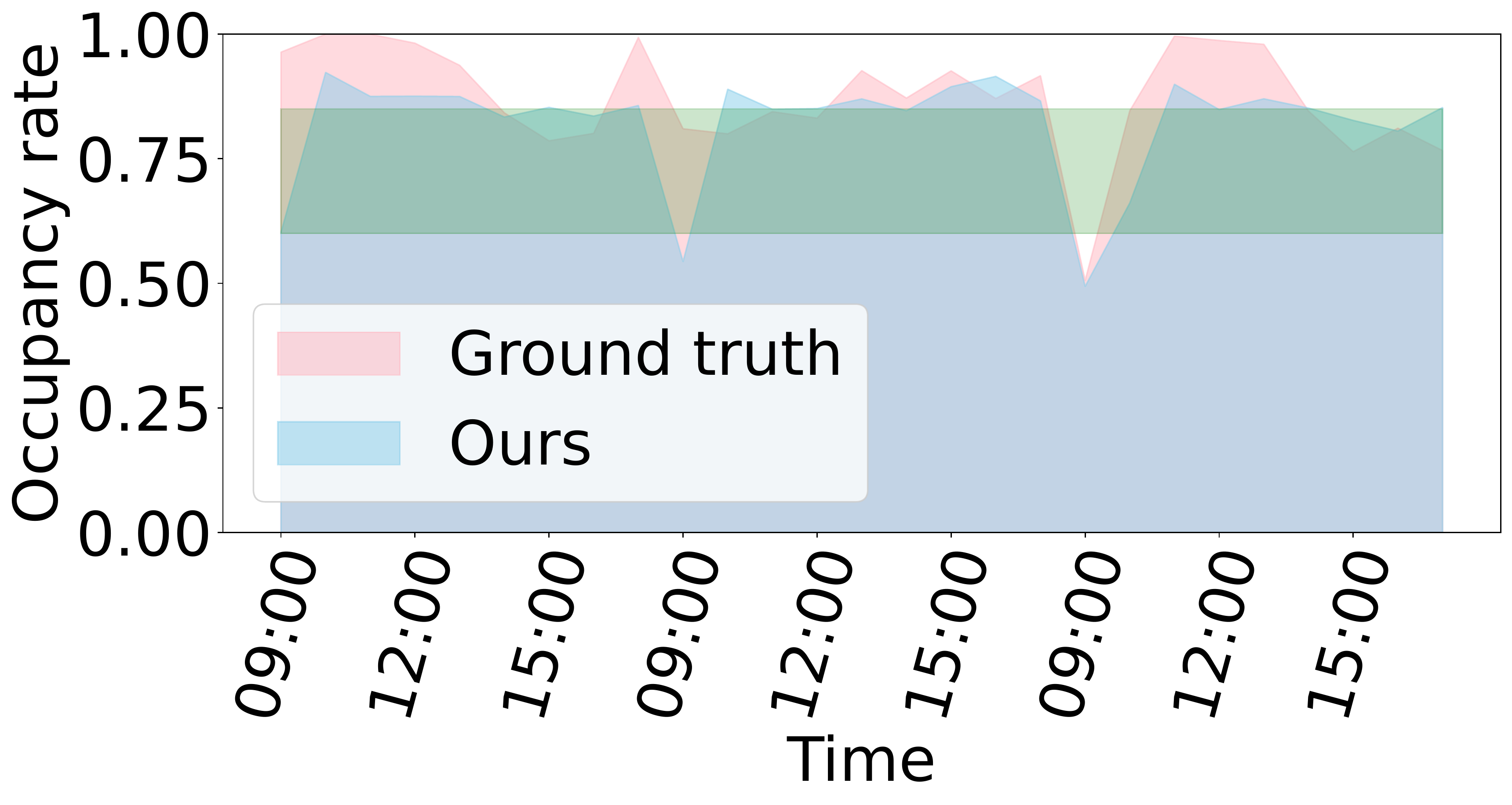}} \\
    \caption{The comparison between the ground-truth occupancy rate and the predicted occupancy rate when our optimized pricing is applied. These figures show Seattle's data.} 
    \label{opt_occ_viz}
\end{figure}

In Fig.~\ref{opt_occ_viz}, we show other types of visualization. In the four parking blocks of Seattle, the ground-truth occupancy rates observed in the data are above the ideal range $[0.6, 0.85]$ in many cases, whereas our method successfully suppresses the occupancy rates on or below the range.

\section{Conclusions \& Limitations}
In the U.S. metropolitan cities, dynamic pricing is necessary to keep parking occupancy rates within a tolerable range. To this end, we presented a one-shot prediction-driven optimization framework which is featured by i) an effective prediction model for the price-occupancy relation, and ii) a one-shot optimization method. Our prediction model is carefully tailored for the price-occupancy relation and therefore, it outperforms other general time series (or spatiotemporal) forecasting models. Our sophisticated prediction model, which includes many layers for processing the short-term and the long-term historical information, and the price information, is designed considering the optimization process, which is quite different from other predictive models. In general, predictive models are not designed considering optimization, and therefore, one should rely on a black-box optimization technique and so on. In our case, however, the price reflection module is deferred to after the initial prediction module, and the final prediction module relies on NODEs which are bijective and continuous. Owing to all these design points, the price optimization can be solved in $\mathcal{O}(1)$. Our experiments with the data collected in San Francisco and Seattle show that the presented dynamic pricing works well as intended, outperforming other optimization methods. Our method is several orders of magnitude faster than them and successfully suppresses too large occupancy rates.

One limitation in our work is that we should have relied on the oracle model to verify the efficacy of our method (instead of running real systems in San Francisco and Seattle). However, this is a common limitation of many prediction-driven optimization researches~\cite{An:2016:MFM:2939672.2939726,An:2017:DFA:3055535.3041217,AAAI18-Chen,li2021largescale,doi:10.1137/1.9781611976700.80,parkape}. We contribute a novel approach, which drastically enhances the complexity of solving the optimization problem, to the community of prediction-driven optimization. 

\section*{Acknowledgement}
Noseong Park is the corresponding author. This work was supported by the Institute of Information \& Communications Technology Planning \& Evaluation (IITP) grant funded by the Korean government (MSIT) (No. 2020-0-01361, Artificial Intelligence Graduate School Program (Yonsei University)).

\clearpage
\bibliographystyle{ACM-Reference-Format}
\balance
\bibliography{ref}

\end{document}